%% file: paper_CAMERAREADY.tex
\documentclass[acmlarge,nonacm]{acmart}

\AtBeginDocument{%
  \providecommand\BibTeX{{%
    \normalfont B\kern-0.5em{\scshape i\kern-0.25em b}\kern-0.8em\TeX}}}

\usepackage{times}
\usepackage{epsfig}
\usepackage{graphicx}
\usepackage{amsmath}
\usepackage{mathtools}
\usepackage[export]{adjustbox} 
\usepackage{booktabs} 
\usepackage{amsfonts}
\usepackage{subfigure}
\usepackage{multirow}
\usepackage{cleveref}
\usepackage{xcolor}
\usepackage{xspace}
\usepackage{dsfont}
\usepackage{eucal}
\usepackage{siunitx}
\crefformat{footnote}{#2\footnotemark[#1]#3}

\usepackage{algorithm}
\usepackage{algpseudocode}
\usepackage{vwcol}  
\usepackage[customcolors,shade]{hf-tikz}

\usepackage{cleveref}
\crefrangelabelformat{figure}%
      {#3#1#4--#5\crefstripprefix{#1}{#2}#6}
\crefrangelabelformat{subfigure}%
      {#3#1#4#5-(\crefstripprefix{#1}{#2}#6}
\Crefname{figure}{Fig.}{Figs.}

\newcommand{\clr}{$\text{CL}^2\text{R}$\xspace}

\newcommand{\neccinformula}{AC\xspace}
\newcommand{\normnecc}{$AC$\xspace}
\newcommand{\lfdm}{$\mathcal{L}_{\scriptscriptstyle \textrm{FD}}^{\scriptscriptstyle \mathcal{M}}$\xspace}
\newcommand{\lfdt}{$\mathcal{L}_{\scriptscriptstyle \textrm{FD}}$\xspace}
\newcommand{\lfdmformula}{\mathcal{L}_{\scriptscriptstyle \textrm{FD}}^{\scriptscriptstyle \mathcal{M}}}

\newcommand{\bc}{$BC$\xspace}
\newcommand{\fc}{$FC$\xspace}

\algdef{SE}[SUBALG]{Indent}{EndIndent}{}{\algorithmicend\ }%
\algtext*{Indent}
\algtext*{EndIndent}

\begin{document}

\title{\clr: Compatible Lifelong Learning Representations}

\author{Niccol\'o Biondi}
\orcid{https://orcid.org/0000-0003-1153-1651}
\affiliation{%
  \institution{University of Florence}
  \streetaddress{V.le Morgagni, 65}
  \city{Florence}
  \country{Italy}
  \postcode{50134}
  }
\email{niccolo.biondi@unifi.it}

\author{Federico Pernici}
\orcid{0000-0001-7036-6655}
\affiliation{%
  \institution{University of Florence}
  \streetaddress{V.le Morgagni, 65}
  \city{Florence}
  \country{Italy}
  \postcode{50134}
  }
\email{federico.pernici@unifi.it}

\author{Matteo Bruni}
\orcid{0000-0003-2017-1061}
\affiliation{%
  \institution{University of Florence}
  \streetaddress{V.le Morgagni, 65}
  \city{Florence}
  \country{Italy}
  \postcode{50134}
  }
\email{matteo.bruni@unifi.it}

\author{Daniele Mugnai}
\orcid{1234-5678-9012}
\affiliation{%
  \institution{University of Florence}
  \streetaddress{V.le Morgagni, 65}
  \city{Florence}
  \country{Italy}
  \postcode{50134}
  }
\email{daniele.mugnai@unifi.it}

\author{Alberto Del Bimbo}
\orcid{0000-0002-1052-8322}
\affiliation{%
  \institution{University of Florence}
  \streetaddress{V.le Morgagni, 65}
  \city{Florence}
  \country{Italy}
  \postcode{50134}
  }
\email{alberto.delbimbo@unifi.it}

\renewcommand{\shortauthors}{Biondi, et al.}

\begin{abstract}
In this paper, we propose a method to partially mimic natural intelligence for the {problem} of \textit{lifelong} learning representations that are \textit{compatible}. We take the perspective of a learning agent that is interested in recognizing object instances in an open dynamic universe in a way in which any update to its internal feature representation does not render the features in the gallery unusable for visual search. We refer to this {learning problem} as \textit{Compatible Lifelong Learning Representations} (\clr) as it considers  {compatible representation} learning {within the} lifelong learning {paradigm}. We identify \textit{stationarity} as the property that the feature representation is required to hold to achieve \textit{compatibility} {and propose a novel training procedure that encourages local and global stationarity on the learned representation}. Due to stationarity, the statistical properties of the learned features do not change over time, making them interoperable with previously learned features. Extensive experiments on standard benchmark datasets show that our $\mathrm{CL^2R}$ training procedure outperforms alternative baselines and state-of-the-art methods. We also provide novel metrics to specifically evaluate compatible representation learning under catastrophic forgetting in {various} sequential learning tasks. Code at \url{https://github.com/NiccoBiondi/CompatibleLifelongRepresentation}.
\end{abstract}



\maketitle

\section{Introduction}
\label{sec:intro}
The universe is dynamic and the emergence of novel data and new knowledge is unavoidable. The unique ability of \textit{natural intelligence} to lifelong learning is highly dependent on memory and knowledge representation \cite{ericsson1995long}. Through memory and knowledge representation, natural intelligent systems continually search, recognize, and learn new objects in an open universe after exposure to one or a few samples. Memory is substantially a cognitive function that encodes, stores, and retrieves knowledge. 
Artificial representations learned by Deep Convolutional Neural Network (DCNN) models \cite{bengio2013representation,taigman2014deepface,deepID_NIPS2014,sharif2014cnn,YosinskiNIPS2014} stored in a memory bank (i.e., the gallery-set) have been shown to be very effective in searching and recognizing objects in an open-set/open-world learning context. Successful examples are face recognition \cite{chopra2005learning,DBLP:conf/cvpr/SchroffKP15,DBLP:conf/cvpr/DengGXZ19}, person re-identification \cite{zheng2015scalable,zheng2016person,zhou2019omni} and image retrieval \cite{aggregating2015yandex, deep2016gordo, tolias2021}.
\begin{figure}[t]
    \centering
    \includegraphics[width=0.7\linewidth]{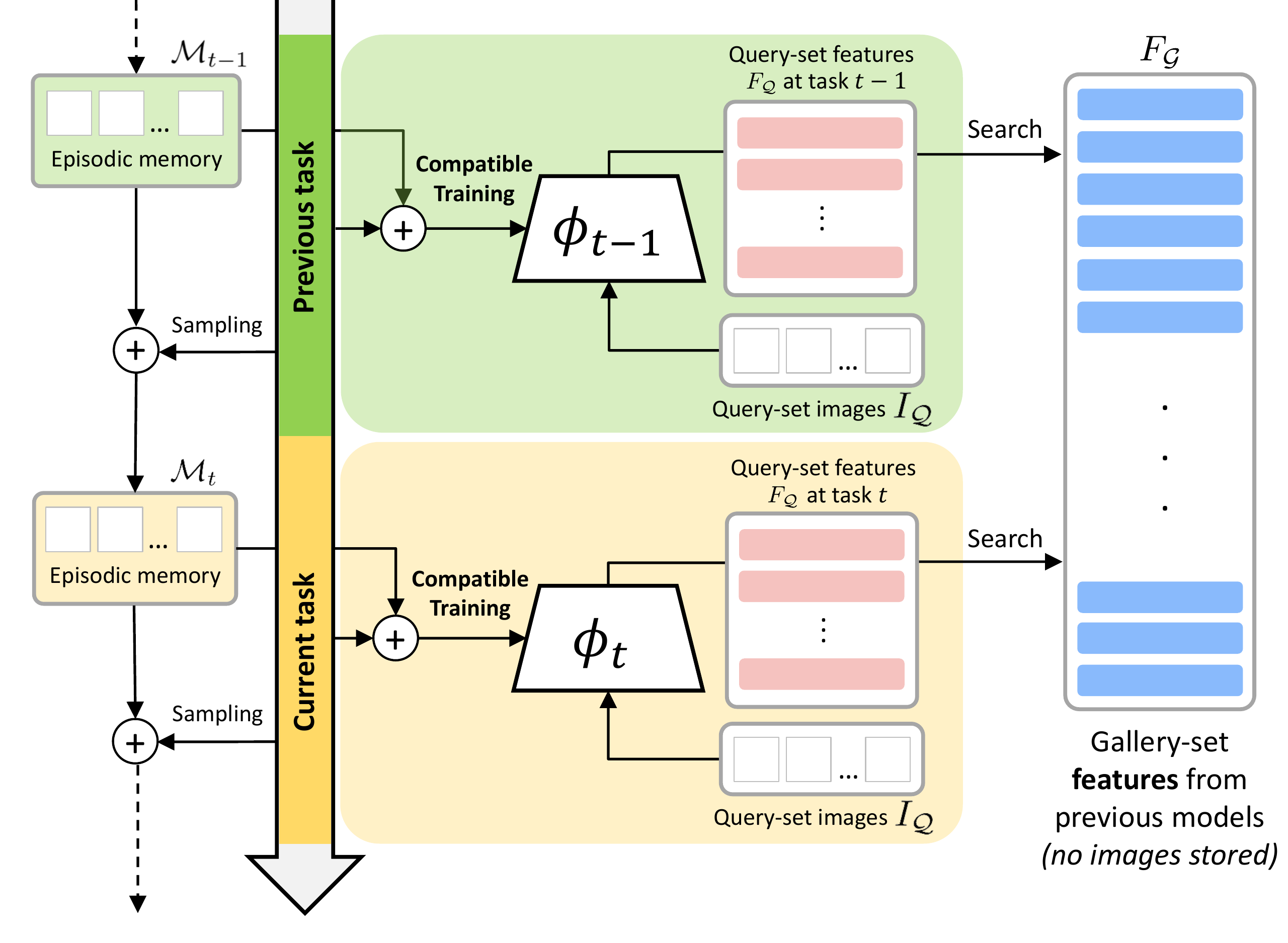}
    \caption{
    {Overview of the Compatible Lifelong Learning Representations (\clr) problem and proposed training procedure.} 
    The learning agent searches object instances from query images $I_\mathcal{Q}$ without re-indexing the gallery-set. Any update to the internal feature representation $\phi$ does not render the features in the gallery-set unusable (i.e., no images are stored). Compatible {feature} representation under catastrophic forgetting is learned imposing stationarity to features learned from the the Class-incremental Learning {surrogate} task. Training is based on rehearsal with the episodic memory $\mathcal{M}_t$.    
}
    \label{fig:cl2r_intro}
\end{figure}
These approaches rely on learning feature representations from static datasets in which all images are accessible at training time. 
On the other hand, dynamic assimilation of new data for lifelong learning suffers from \textit{catastrophic forgetting}: the tendency of neural networks to abruptly forget previously learned information \cite{mccloskey1989catastrophic,ratcliff1990connectionist}.

In the case of visual search, even avoiding catastrophic forgetting by repeatedly training DCNN models on both old and new data, the feature representation still irreversibly \textit{changes} \cite{li2015convergent}. Thus, in order to benefit from the newly learned model, features stored in the gallery must be reprocessed and the ``old'' features replaced with the ``new'' ones. 
Reprocessing not only requires the storage of the original images (a noticeable leap from natural intelligence), but also their authorization to access them \cite{van2020ethical}.
{More importantly, extracting new features at each update of the model} is computationally expensive or infeasible in the case of large gallery-sets. 
The speed at which the representation is updated to benefit from the newly learned data may impose time constraints on the re-indexing process. This may occur from timescales on the order of weeks/months as in retrieval systems or social networks \cite{bct}, to within seconds as in autonomous robotics or real-time surveillance  \cite{Pernici_2018_CVPR,pernici2020self}.
Recently in \cite{bct}, a novel training procedure has been proposed to avoid re-indexing the gallery-set. 
The representation obtained in this manner is said to be compatible, as the features before and after the learning upgrade can be directly compared. Training takes advantage of all the data from previous tasks (i.e., no lifelong learning), guaranteeing the absence of catastrophic forgetting.
The advantage of considering compatible representation learning within the lifelong learning paradigm, as in this paper, is that compatible representation allows visual search systems not only to distribute the computation over time, but also to avoid or possibly limit the storage of images on private servers for gallery data.  
This can have important implications for the societal debate related to privacy, ethical and sustainable issues (e.g., carbon footprint) of modern AI systems~\cite{price2019privacy,van2020ethical,schwartz2020green,cossu2021sustainable}.

We identify \textit{stationarity} as the key requirement for feature representation to be \textit{compatible} during lifelong learning. 
Stationary features have been shown to be biologically plausible in many studies of working memory in the prefrontal cortex of macaques \cite{murray2017stable,meyers2018dynamic,libby2021rotational}. The works \cite{murray2017stable,meyers2018dynamic} decoded the information from the neural activity of the working memory using a classifier with a single \textit{fixed set} of weights. They noted that a \textit{non}-stationary feature representation seems to be biologically problematic since it would imply that the synaptic weights would have to change continuously for the information to be continuously available in memory. 

Inspired by this, in this paper, we formalize the problem of Compatible Lifelong Learning Representations (\clr) in relation to the relevant areas of \textit{compatible learning} and \textit{lifelong (continual) learning}.
We call any training procedure that aims to obtain compatible features and minimize catastrophic forgetting as \clr training, and we propose (1) a novel set of metrics to properly evaluate \clr training procedures (2) a training procedure based on rehearsal \cite{ratcliff1990connectionist,robins1993catastrophic} and feature stationarity \cite{pernici2021regular,pernici2021class} to jointly address catastrophic forgetting and feature compatibility. 
Fig.~\ref{fig:cl2r_intro} provides an overview of the problem and the training procedure. 
Specifically, our \clr training procedure is achieved by encouraging \textit{global} and \textit{local} stationarity to the learned features. 

The paper is organized as follows. In Sec.~\ref{sec:rel_work}, we discuss related work and in Sec.~\ref{sec:main_contrib}, we highlight our contributions. Sec.~\ref{sec:prob} presents the formulation of \clr, Sec.~\ref{sec:compatibility} proposes new metrics to evaluate compatibility, and Sec.~\ref{sec:clr} describes a new training procedure. In Sec.~\ref{sec:results}, we compare our results with adapted state-of-the-art methods. Sec.~\ref{sec:ablation} presents the ablation study.

\section{Related Work} \label{sec:rel_work}

\noindent
\paragraph{\textbf{Compatible Learning.}} 
The work proposed in \cite{bct}, called Backward-Compatible Training (BCT), first formalizes the problem of learning compatible representation to avoid re-indexing. 
The method takes advantage of an influence loss that encourages the feature representation towards one that can be used by the old classifier. The old classifier is fixed while learning with the novel data (i.e., its parameters are no longer updated by back-propagation) and cooperates with the new representation model. Cooperation is achieved by aligning the prototypes of the new classifier with the prototypes of the old fixed one. The underlying assumption is that the upgraded feature representation follows the representation learned by the old classifier.  
BCT has been evaluated in scenarios without the effects of catastrophic forgetting by repeatedly training DCNN models on both old and new images (i.e., jointly re-training from scratch at each upgrade). To compare with this learning strategy in a lifelong learning scenario instead of starting from scratch every time, we have added to BCT the capability of learning by fine-tuning the previously learned model according to a memory based rehearsal strategy \cite{ratcliff1990connectionist,robins1993catastrophic}.

Compatibility under catastrophic forgetting has been implicitly studied in \cite{iscen2020memory} (FAN), in which authors presented a method for storing features instead of images in Class-incremental Learning (CiL). 
They introduce a feature adaptation function
to update the \textit{preserved} features as the network learns novel classes.
We compared to this method by storing the updated-preserved features obtained at each task. Although designed to improve classification accuracy, the work can be considered close to a lifelong learning approach with compatible representation in which the feature adaptation function they defined, addresses implicitly the problem of feature compatibility as in \cite{Chen_2019_CVPR}, \cite{hu2019towards},  \cite{wang2020unified}, \cite{Meng_2021_ICCV}.
Differently from BCT, these methods do not completely prevents the cost of re-indexing since the learned mappings require to be evaluated every time the dataset is upgraded and are therefore not suited to lifelong learning and/or large gallery-set. For example, the ResNet-101 architecture is one order slower than the mapping proposed in \cite{Chen_2019_CVPR}, therefore, when the size of the gallery increases by an order of magnitude, it is equivalent to re-index the images.
The method described in \cite{ramanujan2021forward}, in addition to the current feature model, trains from the same data an auxiliary model in a different way (i.e., using self-supervised learning). The auxiliary feature model will then be used with future learned models to learn a mapping model to obtain compatible representations as in \citep{wang2020unified,iscen2020memory,Meng_2021_ICCV}. The underlying assumption is that as the auxiliary feature model is trained with a different strategy, it encodes different knowledge which may facilitate learning the mapping between the representation spaces. 

Compatibility of the representation in a more general sense has been considered in \cite{li2015convergent,NEURIPS2018_5fc34ed3} where similarity between features extracted from identical architectures and trained from different initialization has been extensively evaluated.
The work in \cite{budnikAsymmetric} avoids re-indexing the gallery although the new model used for queries is not trained on more data. Their work is motivated by the scenario where the gallery is indexed by a large model and the queries are captured from mobile devices in which the use of small models is the only viable solution.

\paragraph{\textbf{Lifelong Learning.}} Lifelong Learning or Continual Learning (CL) studies the problem of learning from a non i.i.d. stream of data with the goal of assimilating new knowledge preventing catastrophic forgetting \cite{lifelongML, mccloskey1989catastrophic}. 
Methods for preventing catastrophic forgetting have been explored primarily in the classification task, where catastrophic forgetting often manifests itself as a significant drop in classification accuracy \cite{vijayan2021continual,delange2021continual,masana2020class,parisi2019continual,belouadah2021comprehensive}.
The key aspects that distinguish lifelong feature learning for visual search from classification are: (i)  categorical data often have coarser granularity than visual search data; (ii) evaluation metrics do not involve classification accuracy; and (iii) class labels are not required to be explicitly learned. These differences may suggest that these two catastrophic forgetting occurrences are of different origins.
In this context, recent works have discussed the importance of the specific task in assessing catastrophic forgetting of learned representations \cite{davari2021probing,chen2020exploration,pu2021lifelong,chen2021feature, barletti2022contrastive,pernici2020icpr}. 
Among others, empirical evidence presented in \cite{davari2021probing} suggests that feature forgetting is not as catastrophic as classification forgetting and that many approaches that address the problem of catastrophic forgetting do not improve feature forgetting in terms of the usefulness of the representation. 
We argue that such evidence is relevant in visual search and that it can be exploited with techniques that further encourage learning compatible feature representation. 
According to this, we consider CiL as the basic building blocks for the general purpose of learning feature representation incrementally.

In this paper, the focus is on CiL methods based on Knowledge Distillation (KD) \cite{hinton2015distilling} and rehearsal \cite{robins1995catastrophic} which are known to be versatile, effective and widely applicable to reduce catastrophic forgetting. We leverage the classification task in CiL as a surrogate task to learn feature representation as typically performed in face/body-identification and retrieval  \cite{DBLP:conf/cvpr/DengGXZ19,zheng2016person,tolias2021}. 
The work in \cite{lwf} first introduces KD in lifelong learning as an effective way to preserve the knowledge previously acquired from old tasks. 
In \textit{iCaRL}~\cite{icarl}, KD is combined with rehearsal to reserve samples of exemplars stored in an episodic memory for classes already seen. 
The BiC work, proposed in \cite{bic}, extends \cite{icarl} developing a bias correction layer to recalibrate the output probabilities learning an additional linear layer on a small set of data. 
Along a similar vein, in \cite{zhao2020maintaining}, the bias correction is performed by aligning the norms of the weight vectors of the classifier for new classes to those for old classes without using additional model parameters or reserved data.
The work in \cite{RomeroBKCGB14} introduces Feature Distillation (FD), a distillation loss evaluated on the feature vectors instead on the classifier outputs. FD has recently been successfully applied in \cite{lucir} (LUCIR) and \cite{douillard2020podnet} to reduce catastrophic forgetting. Differently from LUCIR, PODNet uses a spatial-based distillation loss to constrain the statistics of intermediate features after each residual block. 
Similar to LUCIR, PODNet and many others paper on Continual/Lifelong learning in literature, our problem formulation takes advantage of the general concept of KD. Differently from these works, our approach is novel in that it considers feature distillation for the dual purpose of learning feature compatibility and mitigating feature forgetting.
The work in \cite{iscen2020memory} (FAN), also discussed in the previous paragraph, combines strategies from \cite{lucir}, \cite{lwf}, \cite{icarl} to learn and preserve previous features. Although the work does not consider the compatibility problem, it is the closest work to our approach. 
Recently, \cite{yan2021dynamically} (DER) has shown an interesting performance improvement in Class-incremental Learning by freezing the previously learned representation and expanding its dimension from a new learnable feature extractor. Despite the clear improvements in classification performance, this has no trivial exploitation in compatible training as the varying dimension across tasks do not allow direct application of nearest neighbor search between models. Features with different dimensions typically require to be projected into a common single space to allow nearest-neighbor to be applied. 
The FOSTER method~\cite{wang2022foster} improves upon DER by addressing this specific problem by transforming the growing dimension of the feature representation with a trainable linear layer that maps the growing feature vector into a fixed dimension.
More in general, CiL methods addressing catastrophic forgetting are in a certain sense related to compatible representation, since forgetting is the change in the feature representation of classifiers that will be learned in the future. We evaluate these methods as baselines to quantify the level of lifelong-compatible representation they intrinsically may have.

\section{Main Contributions}\label{sec:main_contrib}

\begin{enumerate}
    
    \item We consider compatible representation learning within the lifelong learning paradigm. We refer to this general learning problem as \textit{Compatible Lifelong Learning Representations} (\clr).
    
    \item We define a novel set of metrics to properly evaluate \clr training procedures.

    \item We propose a \clr training procedure that imposes global and local stationarity on the learned features to achieve compatibility between representations under catastrophic forgetting. {Global and local interactions show a significant performance improvement when local stationarity is promoted only from already observed samples in the episodic memory.}

    \item We empirically assess the effectiveness of our approach in several benchmarks showing improvements over baselines and adapted state-of-the-art methods.
    
\end{enumerate}

\section{\clr Problem Formulation}  \label{sec:prob}
In a \clr setting, a sequence of representation models, $\{ \phi_t \}_{t=1}^{T}$, is learned incrementally with a sequence of $T$ tasks, $\{ (\mathcal{D}_t, K_t) \} _{t=1}^T$, 
where $\mathcal{D}_t$ are the images of the $t$-th task represented by $K_t$ different classes. Specifically, each task is disjoint from the others: $K_k \cap K_t= \emptyset$ with $t \neq k $.
The learned representation model $\phi_t$ is used to transform the query images into feature vectors that are used to retrieve the images most similar to a set of given gallery images transformed with a previous model $\phi_k$. Specifically, we indicate with the couple $\mathcal{G}=(I_\mathcal{G},F_\mathcal{G})$ 
the gallery-set, where ${I}_\mathcal{G}=\{\mathbf{x}_i\}_{i=1}^N$ is the image collection from which the features $F_\mathcal{G}=\{ \mathbf{f}_i \}_{i=1}^N$ are extracted, and $N$ is the number of elements of the two sets. Without loss of generality, we assume that the features in $F_\mathcal{G}$ are extracted using the representation model $\phi_{ k}:{\mathbb R}^D \rightarrow {\mathbb R}^d$ that transforms an image $\mathbf{x} \in {\mathbb R}^D$ into a feature vector $\mathbf{f} \in {\mathbb R}^d$, where $d$ and $D$ are the dimensionality of the feature and the image space, respectively. 
Analogously, we will refer to $\mathcal{Q}=(I_\mathcal{Q},F_\mathcal{Q})$ as the query-set, where $I_\mathcal{Q}$ and $F_\mathcal{Q}$ are the corresponding image-set and the feature-set, respectively.   
As the $t$-th task becomes available, the model $\phi_{t}$ is incrementally learned from the previous one along with the new task data $\mathcal{D}_t$.
Our goal is to design a training procedure to learn the model $\phi_{t}$ so that any query image transformed with it can be used to perform {visual search} through some distance ${\rm dist}:{\mathbb R}^d \times {\mathbb R}^d \rightarrow \mathbb{R}_+$ to identify the closest features ${F}_\mathcal{G}$ to the query features ${F}_\mathcal{Q}$ without forgetting the previous representation and  \textit{without} computing 
\mbox{ $F_\mathcal{G}=\{ \mathbf{f} \in \mathbb{R}^d \, | \,  \mathbf{f} = \phi_{t}(\mathbf{x}) \, \forall \mathbf{x} \in I_\mathcal{G}\}$ (i.e., re-indexing). }
If this holds, then the resulting representation $\phi_{t}$ is said to be \textit{lifelong-compatible} with $\phi_{k}$.

The main challenge of the \clr problem is to jointly alleviate catastrophic forgetting and learn a compatible representation between the previously learned models.
In Fig.~\ref{fig:cl2r_intro} we illustrate the complete \clr training example using rehearsal to alleviate the effects of catastrophic forgetting.

\section{Compatibility Evaluation} \label{sec:compatibility}
A representation model $\phi_{\rm new}$ upgraded with new data is said to be compatible with an old representation model $\phi_{\rm old}$ when it holds~\cite{bct}:
\begin{equation} \label{eq:compatible_set}
M\big(\phi_{\rm new}^{\mathcal{Q}}, \phi_{\rm old}^{\mathcal{G}} \big) > {M} \big(\phi_{\rm old}^{\mathcal{Q}}, \phi_{\rm old}^{\mathcal{G}} \big).
\end{equation}
Eq.~\ref{eq:compatible_set} represents the \textit{Empirical Compatibility Criterion} (ECC), where ${M}$ is an evaluation metric specific to the given visual search problem. 
Notable examples of the metric $M$ can be found in face verification accuracy \cite{LFWTech,LFWTechUpdate}, face verification/identification accuracy in terms of true acceptance and false acceptance rate (TAR$@$FAR) \cite{klare2015pushing}, person re-identification mean average precision (mAP) \cite{ye2021deep}. The intuition of these metrics is based on the observation that they can be instantiated with two different representation models $\phi_{\rm new}$ and $\phi_{\rm old}$ when considering the query-gallery pair.   
The specific notation ${M} \big(\phi_{\rm new}^{\mathcal{Q}}, \phi_{\rm old}^{\mathcal{G}} \big)$ defines the \emph{cross-test} between the new and the old model and it represents the case in which $\phi_{\rm new}$ is used to extract the {features of the query-set, $F_\mathcal{Q}$}, while $\phi_{\rm old}$ is used to extract the gallery-set {ones, $F_\mathcal{G}$}. 
${M} \big(\phi_{\rm old}^{\mathcal{Q}}, \phi_{\rm old}^\mathcal{G} \big)$ is the \textit{self-test} and it represents the case in which both query and gallery features are extracted with $\phi_{\rm old}$. 
When the model is trained incrementally on $T$ tasks, Eq.~\ref{eq:compatible_set} is evaluated according to the Multi-model Empirical Compatibility Criterion introduced in \cite{cores}: 
\begin{eqnarray} \label{eq:multistepecc}
M \big( \phi_t^\mathcal{Q}, \phi_k^\mathcal{G} \big) > 
M \big( \phi_k^\mathcal{Q}, \phi_k^\mathcal{G} \big) {\rm \quad with \:} t > k,
\end{eqnarray}
where $t, k \in \{1,2,\ldots,T\}$ refer to two different tasks such that task $k$ is processed by the model before task $t$. 
The model $\phi_t$ is compatible with the model $\phi_k$, when the \textit{cross-test} $M \big( \phi_t^\mathcal{Q}, \phi_k^\mathcal{G} \big)$ between $\phi_t$ and $\phi_k$ is greater than the \textit{self-test} $M \big( \phi_k^\mathcal{Q}, \phi_k^\mathcal{G} \big)$ of the model $\phi_k$.
The underlying intuition is that if the performance of matching the gallery feature vectors extracted with the old model with the query feature vectors extracted with the new model (i.e., cross-test) is better than the performance of matching the gallery feature vectors with the query feature vectors both extracted with the old model (i.e., self-test), then the system is learning compatible representations. That is,  learning from the new task data improves the representation without breaking the compatibility with the previously learned model.
Based on Eq.~\ref{eq:multistepecc}, the Compatibility Matrix $C$ is defined as:
\begin{equation}
        C_{t, k} =
        \begin{cases*}
          M \big( \phi_t^\mathcal{Q}, \phi_k^\mathcal{G} \big) & if $t > k$ \\
          M \big( \phi_k^\mathcal{Q}, \phi_k^\mathcal{G} \big) & if $t = k$ \\
          \qquad 0 & if $t < k$
        \end{cases*},
        \label{eq:compatibility_matrix}
\end{equation}   
where the element in the row $t$ and the column $k$ of the compatibility matrix denotes the evaluation metric $M$ of the model $t$ to the model $k$. 
{This definition combines the basic intuition of the classification accuracy matrix $R$ defined in \cite{lopez2017gradient, diaz2018don}, used to evaluate the Class-incremental Learning problem (CiL), with the two specific aspects that distinguish the $\text{CL}^2\text{R}$ learning setting from the CiL one. Namely: (a) in CiL at each task, the train and test data are sampled from the same distribution, while in $\text{CL}^2\text{R}$ the test-set classes are sampled from an unknown distribution (i.e., $\text{CL}^2\text{R}$ addresses the open-set recognition problem); (b) in CiL the test-set is dynamic (i.e., it grows including images from the task distributions) while in $\text{CL}^2\text{R}$ it is assumed static for the purpose of a reliable evaluation
\cite{bct}.
In the $\text{CL}^2\text{R}$ setting, a dynamic test-set, as used in CiL, is of difficult definition, as there are infinite ways to make the gallery dynamic and each of them may change unexpectedly the performance of the evaluation. We follow \cite{bct} and perform the evaluation assuming a static test-set (i.e., a static query-gallery pair). 
According to this, we set the elements of the matrix $C$ with $t<k$ to zero to indicate the impossibility of a reliable evaluation of a growing test-set that should be sampled from an unknown changing distribution.
}
For the remaining elements, the cross-test values are the elements of the matrix with $t > k$, while the self-test are the values of the main diagonal (i.e., when $t = k$). Given a compatibility matrix $C$, the Average Compatibility (\normnecc)
is defined as follows:
\begin{equation}
\neccinformula = \frac{2}{T(T-1)} \sum\limits_{1 \leq k < t \leq T}\mathds{1}{ \Big (M \big( \phi_t^\mathcal{Q}, \phi_k^\mathcal{G} \big) >  M \big( \phi_k^\mathcal{Q}, \phi_k^\mathcal{G} \big)} \Big ),
\label{eq:norm_mecc}
\end{equation}
where $\mathds{1}(\cdot)$ denotes the indicator function. The Average Compatibility (\normnecc) summarizes the Compatibility Matrix values in a single number that quantifies the number of times that compatibility is verified against all possible $\frac{T(T-1)}{2}$ occurrences.

\subsection{Proposed \clr Metrics} \label{sec:metrics}
The work in \cite{lopez2017gradient, diaz2018don} proposes a set of metrics to assess the ability of the learner to transfer knowledge based on a matrix that reports the test classification accuracy of the model on task $j$ after learning task $i$.
Along a similar vein, we present a set of metrics to evaluate the compatibility between representation models in a compatible lifelong learning setting.

Let $C \in \mathbb{R}^{ T \times T}$ be the compatibility matrix of the Eq.~\ref{eq:compatibility_matrix} for $T$ tasks, the proposed criteria are the following:
\begin{enumerate}
    \item \emph{Backward Compatibility} (\bc) 
    measures the gap in compatibility performance between the representation learned at task $T$ with respect to the representation learned at task $k$ with $k \in \{1, \ldots, T-1\}$.
    When \textit{\bc} $< 0$ the learning procedure is also influenced by catastrophic forgetting because the performance degrades with newer learned tasks. $BC$ is defined as follows:
    \begin{equation}
        \mbox{\normalsize \bc} = \frac{1}{T-1} \sum_{k=1}^{T-1} \big ( C_{T,k} - C_{k,k}  \big ) 
        \label{eq:back_comp}
    \end{equation}
    
    \item \emph{Forward Compatibility} (\fc)
    estimates the influence that learning a representation on a task $k-1$ has on the compatibility performance of the representation learned on a future task $k$ by comparing the cross-test (between models at task $k$ and $k-1$) with respect to the self-test at task $k$.
    \fc $\ge 0$ denotes that, on average, the cross-test values are greater than the self-test evaluated on the subsequent tasks, therefore, re-indexing does not necessarily provide improved results.
    \fc is defined as follows:
    \begin{equation}
        \mbox{\normalsize \fc} = \frac{1}{T-1} \sum_{k=2}^{T} \big ( C_{k,k-1} - C_{k,k} \big ).
        \label{eq:forw_comp}
    \end{equation}

    The intuition behind the definition of this metric comes from noticing that as the number of tasks increases, the \textit{cross-test} may result better than the \textit{self-test}. As this is not typically observed when there is no catastrophic forgetting (i.e., when repeatedly training with new and old data),
    we argue this is due to the \textit{joint interaction} between the compatibility constraint and catastrophic forgetting. This observation led us to define something ``positive'' when the compatible representation with the previously learned model is higher than the self-test of the current model. This metric is designed to yield high values when a \clr training procedure is able to positively exploit the joint interaction between feature forgetting and compatible representation.
    
\end{enumerate}
\noindent
From Eqs.~\ref{eq:back_comp} and \ref{eq:forw_comp}, it can be deduced that \bc and \fc $\in [-1,1]$. 
Backward compatibility for the first task and forward compatibility for the last task are not defined.
The larger these metrics, the better the model. 
When \normnecc values are comparable, both \bc and \fc represent two metrics that quantify the positive interaction between search accuracy under catastrophic forgetting and compatibility. This allows evaluating how catastrophic forgetting affects the representation and its compatibility.

As \bc evaluates the relationship between the representations learned at the final task $T$ and the previous ones, it is possible to follow their evolution during \clr training. 
According to this, we define the Backward Compatibility at task $t$ as $BC{(t)} = \frac{1}{t-1} \sum_{\substack{k=1}}^{t-1} \big( C_{t,k} - C_{k,k} \big), \; {\rm with  } \; t > 1$
where $t \in \{1, 2, \ldots, T\}$.
This represents the average of the element-wise difference between the $t$-th row and the first $t$ elements of the main diagonal on the compatibility matrix.

\section{Proposed \clr Training} 
\label{sec:clr} 
To achieve compatibility, we encourage \textit{global} and \textit{local} stationarity to the feature representation.
\noindent
Global stationarity is encouraged according to the approach described in \cite{pernici2021regular}, in which features are learned to follow a set of special fixed classifier prototypes. 
The work \cite{pernici2021regular} imposes global stationarity using a classifier in which prototypes cannot be trained (i.e., fixed) and are set before training. Under this condition, only the direction of the features aligns toward the fixed directions of the classifier prototypes and not the opposite. This constraint imposes learned features to follow their corresponding fixed prototypes, therefore, encouraging representation stationarity. The lack of trainable classifier functionality is basically replaced by previous layers. Fixed prototypes are set according to the coordinate vertices of a $d$-Simplex regular polytope that, in addition to stationarity, allows maximally separated features to be learned \cite{pernici2019fix, Pernici_2019_CVPR_Workshops}. 

We take advantage of this result and perform Class-incremental Learning (CiL) as surrogate task to learn stationary features' representation to achieve compatibility.
More formally, let $ \mathbf{W} \; \forall t \in \{1, 2, ..., T\}$ be the $d$-Simplex fixed classifier, we instantiate the CiL problem as $\sigma(\phi_t \circ \mathbf{W})$, where $\sigma$ indicates the softmax function, and perform learning according to incremental fine-tuning. The evolving training-set $\mathcal{T}_t \gets \mathcal{M}_{t} \cup \mathcal{D}_t$ is computed according to a rehearsal base strategy using the episodic memory, $\mathcal{M}_{t}$ which contains an updating set of samples from $\{\mathcal{D}_1, ..., \mathcal{D}_{t-1} \}$. The memory is updated as $\mathcal{M}_{t+1} \gets \mathcal{M}_{t} \cup \textsc{Sampling}({\rm }D_{t})$.  
The loss optimized in \cite{pernici2021regular} is adapted to \clr training as follows:
\makeatletter
\newcommand\norm[1]{\left\lVert#1\right\rVert}
\newcommand{\mvast}{\bBigg@{2.5}}
\newcommand{\vast}{\bBigg@{3.5}}
\newcommand{\Vast}{\bBigg@{5}}
\makeatother

\begin{eqnarray}
    \mathcal{L}_t=
    -\dfrac{1}{|\mathcal{T}_{t}|} 
    \sum\limits_{\mathbf{x} \in \mathcal{T}_{t}} \log \! 
    \Bigg(
        \dfrac {\exp{ \big( {\mathbf{w}}_{y_i}^{\top}\cdot{\mathbf{\phi(\mathbf{x})}} }\big)} {\sum\limits_{\scriptscriptstyle j \in K_s} \exp\big({  {\mathbf{w}}_{j}^{\top}\cdot{\mathbf{\phi(\mathbf{x})}} }\big) + \sum\limits_{\scriptscriptstyle j \in K_u} \exp{\big(  {\mathbf{w}}_{j}^{\top}\cdot{\mathbf{\phi(\mathbf{x})}} \big) }}
    \Bigg)
    \label{eq:loss_ce},
\end{eqnarray}
where $K_s$ is the set of classes learned up to time $t$, $|\mathcal{T}_{t}|$ is the number of elements in the training-set, $K_u$ is the set of the outputs of the classifier that have not yet been assigned to classes at time $t$ (i.e., future unseen classes \cite{pernici2021class}), $\mathbf{w}^{\top}_{\resizebox{0.01\hsize}{!}{$(\cdot)$}}$ is a class prototype of the fixed classifier $\mathbf{W}$, and $y_i$ is the supervising label.
In particular, $\mathbf{W}$ is the weight matrix of the fixed classifier, which does not undergo learning during model training. 
In \cite{pernici2021regular} the $d$-Simplex prototypes are defined as $\mathbf{W} = \{e_1,e_2,\dots,e_{d-1}, \alpha \sum_{i=1}^{d-1} e_i \},$ where $d$ the feature dimensionality of the $d$-Simplex, $\alpha=\frac{1-\sqrt{d+1}}{d}$, and $e_i$ denotes the standard basis in $\mathbb{R}^{d-1}$, with $i \in \{1,2, \dots, d-1\}$. 

The loss of Eq.~\ref{eq:loss_ce} imposes global stationarity and does not require any knowledge to be extracted from the previously learned models. However, catastrophic forgetting causes misalignment between features and fixed classifier prototypes. 
Therefore, we further impose additional stationarity constraints in a local neighborhood of a feature by encouraging the current model to mimic the feature representation of the model previously learned. This allows the overall stationarity to also be determined by a local learning mechanism interacting with the global one provided by the $d$-Simplex classifier of Eq.~\ref{eq:loss_ce}. The global-to-local interaction is achieved through the Feature Distillation loss (FD) \cite{RomeroBKCGB14}. Differently from the more common practice of FD in CiL \cite{less_forgetful, lucir, douillard2020podnet} in which each mini-batch is sampled from both the episodic memory and the current task (i.e., $\mathcal{T}_t \gets \mathcal{M}_{t} \cup \mathcal{D}_t$), we evaluate the FD loss, at each task $t$, \textit{only} on the samples stored in episodic memory $\mathcal{M}_t$ observed from previous tasks: 
\begin{equation}
    \label{eq:feat_dist_on_mem}
    \lfdmformula = \frac{1}{|\mathcal{M}_{t}|} \sum_{\mathbf{x}_i \in \mathcal{M}_{t}} \mvast( 1 - \frac{\phi_{t}(\mathbf{x}_i) \cdot \phi_{t-1}(\mathbf{x}_i)}{\norm{\phi_{t}(\mathbf{x}_i)}  \norm{\phi_{t-1}(\mathbf{x}_i)}}  \mvast),
\end{equation}
where $\phi_{t-1}$ is the model learned from the previous task.
This encourages local stationarity and stability from only the previous classes in the episodic memory and the assimilation of new knowledge (plasticity) from only the classes of the current task.
As confirmed by ablation in Sec.~\ref{sec:ablation}, this learning strategy leads to a significant performance improvement.
The final optimized loss function is the sum of Eq.~\ref{eq:feat_dist_on_mem} and Eq.~\ref{eq:loss_ce} 
\begin{equation}
\label{eq:loss_clr}
    \mathcal{L} = \mathcal{L}_{t} + \lambda \; \lfdmformula,
\end{equation}
where $\lambda$ balances the contribution of global and local alignment provided by the two losses. The pseudo-code in Algorithm~\ref{alg:cl2r_train} and in Algorithm~\ref{alg:cl2r_search} detail our training procedure and its application in visual search, respectively.

\input{algorithm/cl2r_procedure}

\section{Experimental Results} \label{sec:results} 
\subsection{Datasets and Verification Protocol}
\label{sec:datasets_task}
We compare our proposed \clr training procedure and the baseline methods on several benchmarks: CIFAR10~\cite{cifar}, ImageNet20\footnote{ To meet the open-set protocol, we generated a training set from ImageNet~\cite{russakovsky2015imagenet} by randomly sampling 20 classes that are not included in the Tiny-ImageNet200 dataset. The indices of the ImageNet classes we use are the following: \{n02276258, n01728572, n03814906, n02817516, n03769881, n03220513, n04442312, n04252225, n13037406, n04266014, n03929855, n02804414, n01873310, n03532672, n01818515, n03916031, n03345487, n02114855, n04589890, n03776460\}.}, ImageNet100~\cite{icarl, lucir, bic}, Labeled Face in the Wild (LFW)~\cite{LFWTech}, and IJB-C~\cite{maze2018iarpa}. 
Evaluation is performed in the open-set 1:1 search problem, with verification accuracy as the performance metric $M$ in Eq.~\ref{eq:compatible_set} and Eq.~\ref{eq:multistepecc} for all datasets except IJB-C in which the true acceptance rate at false acceptance rate (TAR@FAR) is used. They are defined as $\text{TAR} = {\text{TP}}/{(\text{TP} + \text{FN})}$, $\text{FAR} = {\text{FP}}/{(\text{FP} + \text{TN})}$ and $\text{ACC} = {(\text{TP} + \text{TN})}/{(\text{TP} + \text{TN} + \text{FP} + \text{FN})}$, where TP, TN, FP, FN indicate true positives, true negatives, false positives, and false negatives, respectively \cite{klare2015pushing,salehi2021unified}. 
Following the verification protocol defined in \cite{LFWTech}, we generate a set of pairs of images that do or do not belong to the same class. 
A pair is verified on the basis of the distance between feature vectors of its samples. During the evaluation of task $t$, $\phi_t$ is used to extract the feature representation for the first image of each pair (i.e., the query-set) and $\phi_k$, with $k \in \{1, \ldots, t\}$, is used to extract the feature representation for the second image (i.e., the gallery-set).
When $k=t$, the compatibility test is the self-test, otherwise it is the cross-test between the two representations learned from the tasks at time $t$ and $k$.
For the LFW and IJB-C evaluation, we use the original pairs provided by the respective datasets; for the CIFAR10, ImageNet20, and ImageNet100 evaluation, the verification pairs are randomly generated.
As the open-set evaluation requires no overlap between classes of the training-set and test-set, we use CIFAR100 to perform Class-incremental Learning (i.e., classification is the surrogate task from which the feature representation is learned) and the CIFAR10 pairs are used as verification test-set. Similarly, Tiny-ImageNet200~\cite{le2015tiny} is used as training-set to evaluate the ImageNet20 pairs; LFW and IJB-C pairs are evaluated with models trained on CASIA-WebFace~\cite{DBLP:journals/corr/YiLLL14a}. Finally, for ImageNet100, we train the models with images not included in ImageNet100, i.e., the subset of the images of the remaining 900 classes that we named ImageNet900.
These datasets are divided into tasks as described in Sec.~\ref{sec:implementation_details}.

\subsection{Implementation Details}\label{sec:implementation_details}
Our \clr training procedure is implemented in PyTorch~\cite{paszke2019pytorch} and uses the publicly available library Continuum~\cite{continuum}. We used 4 NVIDIA Tesla A100 to train the representation models, the neural network architectures are based on the PODNet implementation\footnote{\label{footnote:doullard} \url{https://github.com/arthurdouillard/incremental_learning.pytorch}}.
The evaluation is carried out on several ResNet \cite{resnet} architectures. 
Specifically, a 32, 18, and 50 layers ResNet is used for: CIFAR10; ImageNet20 and ImageNet100; and for LFW and IJB-C, respectively.  
As typically used in CiL \cite{lucir, bic}, the episodic memory $\mathcal{M}$ contains 20 samples for each class. 
The value of $\lambda$ in Eq.~\ref{eq:loss_clr} is set $\lambda = \lambda_{\rm base} \sqrt{{k_n}/{k_0}}$ \cite{lucir}, 
in which $\lambda_{\rm base}$ is a scalar, $k_n$ is the number of classes of the current task and $k_0$ is the number of old classes in the episodic memory.
The training details for each dataset are listed below.

\noindent 
\textbf{CIFAR100 and CIFAR10.} We train the model for 70 epochs for each task with batch size 128, optimization is performed with SGD with an initial learning rate of 0.1 and weight decay of $ 2\cdot10^{-4} $. The learning rate is divided by 10 at epochs 50 and 64. 
The input images are RGB, $32 \times 32$.  $\lambda_{\rm base}$ is set to 5.

\noindent 
\textbf{Tiny-ImageNet200 and ImageNet20.} We train the model for 90 epochs at each task with batch size 256, optimization is performed with SGD with an initial learning rate of 0.1 and a weight decay of $ 2\cdot10^{-4} $. The learning rate is divided by 10 at epochs 30 and 60.
To properly evaluate the models in this learning setting, input images and the ImageNet test images are resized to match the Tiny-ImageNet200 input size (RGB $64 \times 64$). 
$\lambda_{\rm base}$ is set to 5.

\noindent 
\textbf{ImageNet900 and ImageNet100.} We train the model for 90 epochs in each task with batch size 256, optimization is performed with SGD with an initial learning rate of 0.1 and weight decay of $ 2\cdot10^{-4} $. The learning rate is divided by 10 at epochs 30 and 60.
The input images are RGB, $224 \times 224$. 
$\lambda_{\rm base}$ is set to 10.

\noindent 
{
\textbf{CASIA-WebFace and LFW/IJB-C.}} For each task, we train the model for 120 epochs with batch size 1024. Optimization is carried out with SGD with an initial learning rate of 0.1 and a weight decay of $ 2\cdot10^{-4} $. The learning rate is divided by 10 at epoch 30, 60 and 90. The input size are RGB images $112 \times 112$.  $\lambda_{\rm base}$ is set to 10.

In Tab.~\ref{tab:datasets}, we summarize the datasets and the training details of our experiments.

\input{tables/datasets}

\subsection{Baselines and Compared Methods}
We compare our training procedure with both the CiL methods and the recently proposed methods for compatible learning. Our baselines include {LwF}~\cite{lwf}, {LUCIR}~\cite{lucir}, {BiC}~\cite{bic}, PODNet~\cite{douillard2020podnet}, FOSTER~\cite{wang2022foster}, {FAN}~\cite{iscen2020memory}, and {BCT}~\cite{bct}. 
In particular, FAN and BCT are the only approaches with an explicit mechanism to address feature compatibility. 
We adapted FAN so that the learned adaptation functions are used to transform the features into compatible features.     
Since in BCT the model is trained from scratch at each task using all available data, for a fair comparison, we also re-implemented it with an episodic memory and refer to it as lifelong-BCT ($\ell$-BCT). 
At each task, the model is initialized with the parameters of the model of the previous task and the data of the previous tasks can be accessed only through the episodic memory.
For LwF, BiC, and PODNet we use their publicly available implementations\cref{footnote:doullard}, while for LUCIR and FOSTER we adopted their official implementations\footnote{\label{impl:lucir} \url{https://github.com/hshustc/CVPR19_Incremental_Learning} and \url{https://github.com/G-U-N/ECCV22-FOSTER}}.
Finally, we also include a traditional experience replay-based baseline, denoted as ER, where the model is continuously fine-tuned as new tasks become available. 
Finally, in order to evaluate our training procedure without considering the catastrophic forgetting phenomenon, we define as upper bound (UB) our training procedure re-trained from scratch at each task using an episodic memory with infinite size.

\subsection{Evaluation on CIFAR10}
\label{sec:eval_cifar10}

\input{tables/cifar/2task_cl2r_baseline}
\input{tables/cifar/3task} 

In this section, we report the experiments in two, three, five, and ten tasks \clr settings with models trained on CIFAR100 (i.e., using 50, 33, 20, 10 classes per task) where compatibility is evaluated on the CIFAR10 generated pairs.

In Tab.~\ref{tab:baselinecomp}, we summarize the performance of our \clr training procedure with respect to the other baselines in the two-task scenario. 
We evaluate the compatibility of the updated model according to the ECC (Eq.~\ref{eq:compatible_set}), \bc (Eq.~\ref{eq:back_comp}), and \fc (Eq.~\ref{eq:forw_comp}).
The first row of Tab.~\ref{tab:baselinecomp} reports the verification accuracy of the model trained on the first 50 classes of CIFAR100. 
Experiments show that, among the methods compared, LUCIR {and PODNet} may have an inherent, although limited, level of compatible representations. This substantially confirms the importance of having some form of mechanism to preserve the local geometry of the learned features. 
Our training procedure achieves the highest cross test, \bc, and \fc, thus resulting to be the most suited training procedure to avoid re-indexing. 

{In the last rows of the table, we report the performance of the BCT and our upper bound (UB) that are not affected by catastrophic forgetting. The effect of catastrophic forgetting and its implications on the reduction of performance in compatibility can be observed in the self-test, as these values are significantly higher than the values reported by the methods learned using CiL.
}

In Tab.~\ref{tab:3step}, results for the scenario of three, five, and ten-task \clr are presented. For each experiment, we report \normnecc (Eq.~\ref{eq:norm_mecc}), \bc (Eq.~\ref{eq:back_comp}), and \fc (Eq.~\ref{eq:forw_comp}).
As can be noticed, our method always achieves the highest \normnecc, thus obtaining the largest number of compatible representations between models, and always achieves the highest \bc between methods that are subject to catastrophic forgetting. 
FAN achieves almost the same performance as our procedure in the three-task scenario, while, when the number of tasks increases, it has a significant decrease in performance, especially in the ten-task setting. This may be due to increasing number of adaptation functions between different feature spaces that FAN uses to adapt old features with respect to the new ones. 
As can be noticed from the two tables FOSTER does not learn compatible features. This may be due to the fact that feature space compression forces the representation to change abruptly reducing the overall compatibility with previous models.
BCT reports higher values since its representation is learned from scratch for each new task.
Compared to the upper bound (UB), our training procedure achieves lower \normnecc and \bc, this is due to the influence of catastrophic forgetting.
From the table it can also be noticed that BiC, LUCIR, {and PODNet} do not satisfy compatibility when catastrophic forgetting is more severe, as, for example, in the case of ten-tasks. 
Overall, these results suggest that the interaction between local and global stationarity promoted by our training procedure shows a significant improvement in performance that feature distillation alone cannot provide.

\input{tables/tiny/2task}
\subsection{Evaluation on ImageNet}
In this section, we conducted the experiments with models trained on Tiny-ImageNet200 in \clr setting with two (Tab.~\ref{tab:two_tiny}), three, five and ten (Tab.~\ref{tab:multi_tiny}) tasks.

Tab.~\ref{tab:two_tiny} follows the same structure as Tab.~\ref{tab:baselinecomp} showing the ECC (Eq.~\ref{eq:compatible_set}), \bc (Eq.~\ref{eq:back_comp}), and \fc (Eq.~\ref{eq:forw_comp}) values. 
For all the compared methods, the initial model (i.e, the previous model) is trained on the first 100 classes of Tiny-ImageNet200. 
As can be seen in the table, our method achieves the best performance. Although with low values, other methods such as FAN and LUCIR have a certain level of compatibility, which confirms again that distillation, with which they are equipped, is a useful tool to support learning compatible features.
As also observed in the CIFAR results, methods not subject to catastrophic forgetting, (i.e., BCT and ours upper bound), achieves higher \bc and lower \fc.

\input{tables/tiny/multi_task}

Tab.~\ref{tab:multi_tiny} shows the three, five, and ten-task \clr settings for Tiny-ImageNet200.
In these learning scenarios, each task is made up of 66, 40 and 20 classes, respectively.
In this table, we discuss the results by analyzing the values of \normnecc (Eq.~\ref{eq:norm_mecc}), \bc (Eq.~\ref{eq:back_comp}), and \fc (Eq.~\ref{eq:forw_comp}). Our approach always achieves the highest value of \normnecc. In particular, ER, LwF, BiC, FAN, and $\ell$-BCT do not achieve lifelong-compatible representation in the three-task setting as a result of \normnecc = 0. 
In the ten-task \clr setting, it is more evident that as the number of tasks increases, methods without any specific mechanism to preserve the representation typically cannot learn compatible representations. LUCIR, BiC, and $\ell$-BCT obtain significantly lower values than our method.
Specifically, the \normnecc performance is more than twice that of BCT, which means that our \clr procedure obtains twice the number of compatible representations than that of BCT. This may be caused by the fact that the constraints imposed by these techniques on the learned representation seem to have very little effect on its stationarity and consequently in its compatibility.
{
The results on the ten task are also important as they suggest that catastrophic forgetting is not an intrinsic impediment to learning compatible representations. 
}
The performance difference of 0.11 in \normnecc with respect to the upper bound can be considered clear evidence of this effect. 
Finally, the table shows how our training procedure provides the highest \fc and is the only case where \fc is always positive.
As a result, our training procedure achieves, on average, cross-tests higher than self-tests indicating that the system performs better even without re-indexing the gallery. 

\input{tables/tiny/imagenet}
Tab.~\ref{tab:imagenet} reports ImageNet100 results when models are trained on ImageNet900 with two and three tasks. We compare our approach with the $\ell$-BCT method as having reasonable performance and with an explicit mechanism to learn compatible features under catastrophic forgetting. As can be noticed from the table, our \clr training clearly outperforms $\ell$ -BCT. Our method achieves good scores for the AC in both scenarios. As remarked in the Sec.7.5 of the novel revised manuscript, the reduced performance of $\ell$-BCT appears to be connected to the fact that the training procedure is only based on pairwise model training (i.e., compatibility is only learned from the previous model). In contrast, our method is not based only on pairwise learning and does not use previous classifiers, which may be incorrectly learned.

\subsection{Face Verification}

In this section, we report the experimental results on the LFW and IJB-C benchmarks in two, three, five and ten \clr settings.
We incrementally train the representation models with CASIA-WebFace resulting in tasks composed of 5287, 3525, 2115, and 1057 classes, respectively.

The results are summarized in Tab.~\ref{tab:face} and Tab.~\ref{tab:face_ijbc-all} for LFW  and IJB-C, respectively. 
In particular, for IJB-C, we report accuracy in terms of \normnecc, \bc and \fc at different false acceptance rates (FAR): $10^{-1}$, $10^{-2}$, { and $10^{-4}$}.
In this evaluation, we do not report LUCIR when training on the CASIA-WebFace due to the excessive memory requirements of the original implementation\cref{impl:lucir}. 
Although in the two-task scenario comparable results are observed to those of $\ell$-BCT, in the settings of three and five tasks, our training procedure achieves complete compatibility resulting in \normnecc = 1 and  \bc  always positive. In ten-task compatibility, the difference in performance increases more significantly, confirming a clear overall positive performance. 
Generally, the reported performances are higher on face datasets than on CIFAR10, ImageNet20, and ImageNet100. Possible reasons may be found in the fact that in face recognition, the domain shift between classes is lower than that for CIFAR or ImageNet. 
Finally, this experiment shows that the proposed method is effective not only with a larger number of model updates, but also with larger datasets. 

\input{tables/face/multi_task}
\input{tables/face/ijbc}
\input{figures/bc_ca_per_task}

\subsection{Compatibility \& Catastrophic Forgetting}
In this section, we study how compatibility is related to the problem of catastrophic forgetting. In Fig.~\ref{fig:metric_per_task}, we show the evolution of \bc in a five and ten-task \clr scenario.
In particular, \Crefrange{fig:bc_per_task_5_cif}{fig:bc_per_task_10_cif} and in \Crefrange{fig:bc_per_task_5_tiny}{fig:bc_per_task_10_tiny} show the evaluations on the CIFAR10 and ImageNet20 datasets, respectively.
We compared our approach with ER, LwF, BiC, LUCIR, FAN and $\ell$-BCT. 
As can be observed, our training procedure achieves the highest performance.  
As the \bc metric is, on average, the closest to zero than the other evaluated methods, the representation learned by our training procedure can be considered to be the most compatible and, from the perspective of visual search, equivalent to the representation models learned from previous tasks. More practically, this allows for the reduction of the computational cost of re-indexing.

In contrast, FAN achieves a negative value of \bc in all four settings, confirming that the composition of an increasing number of feature adaption functions between sequentially learned representations causes a decrease in compatibility. Despite the absence of considerable performance loss, as in the case of FAN, negative \bc values indicate a constant deterioration in performance as the number of tasks increases.
In general, except for our method, the figure shows that all the other methods follow a common trend with lower performance.

\section{Ablation Studies} \label{sec:ablation}
We analyze by ablation the main components of our training procedure. The ablation is performed on the CIFAR100 dataset as described in Sec~\ref{sec:eval_cifar10} and considers the ten-task \clr setting which can be regarded as a worst-case scenario for this dataset. We analyze the impact of: (a) the specific classifier: Trainable vs. Fixed $d$-Simplex with or without the feature distillation component, (b) how the FD loss is evaluated, and (c) the sensitivity of the number of samples reserved per class in the episodic memory.

\input{tables/cifar/ablation_repo-fd}
\paragraph{\textbf{Impact of the $d$-Simplex fixed classifier and Feature Distillation}}
As it can be noticed from Tab.~\ref{tab:abl_repo-fd}, the Trainable classifier is not able to learn compatible representations. When combined with feature distillation, the performance improves only marginally and not sufficiently to be compared with the CiL approaches shown in Tab.~\ref{tab:3step}.
Feature Distillation evaluated on the only samples stored in episodic memory as defined in \lfdm (Eq.~\ref{eq:feat_dist_on_mem}), improves the values of the reported metrics showing a better supervision signal for the updated model. 
The $d$-Simplex alone improves on the previous components obtaining a values of \normnecc=$0.27$ and \fc=$0.003$, which are higher than the Trainable classifier with \lfdm. This remarks on the importance of preserving the global geometry of the learned features according to the $d$-Simplex fixed classifier. 

\paragraph{\textbf{Impact of Memory Samples on Feature Distillation}}
Tab.~\ref{tab:abl_repo-fd} shows that when the distillation loss is evaluated on the only samples stored in episodic memory \lfdm (Eq.~\ref{eq:feat_dist_on_mem}), our approach achieves the better overall results.
We argue this positive effect is mostly due to the interaction between the global feature stationarity learned using the fixed classifier and the local one promoted through feature distillation from the only observed samples in the episodic memory. The interaction is most likely related to the fact that the fixed $d$-Simplex classifier in general does not allow novel classes to interfere in the feature space of the already learned one. 
This in turn provides favorable working conditions (i.e., a kind of coarse pre-alignment) for achieving feature alignment with respect to the previous model by the distillation loss. As expected, the impact intensifies when evaluated only on  already known classes, as alignment is less prone to unexpected noisy features which may reduce the degree of the alignment. This confirms the effectiveness of restricting FD only on memory samples in contrast to the traditional feature distillation commonly used in CiL.
\begin{figure}[t]
    \centering
    \includegraphics[width=0.7\linewidth]{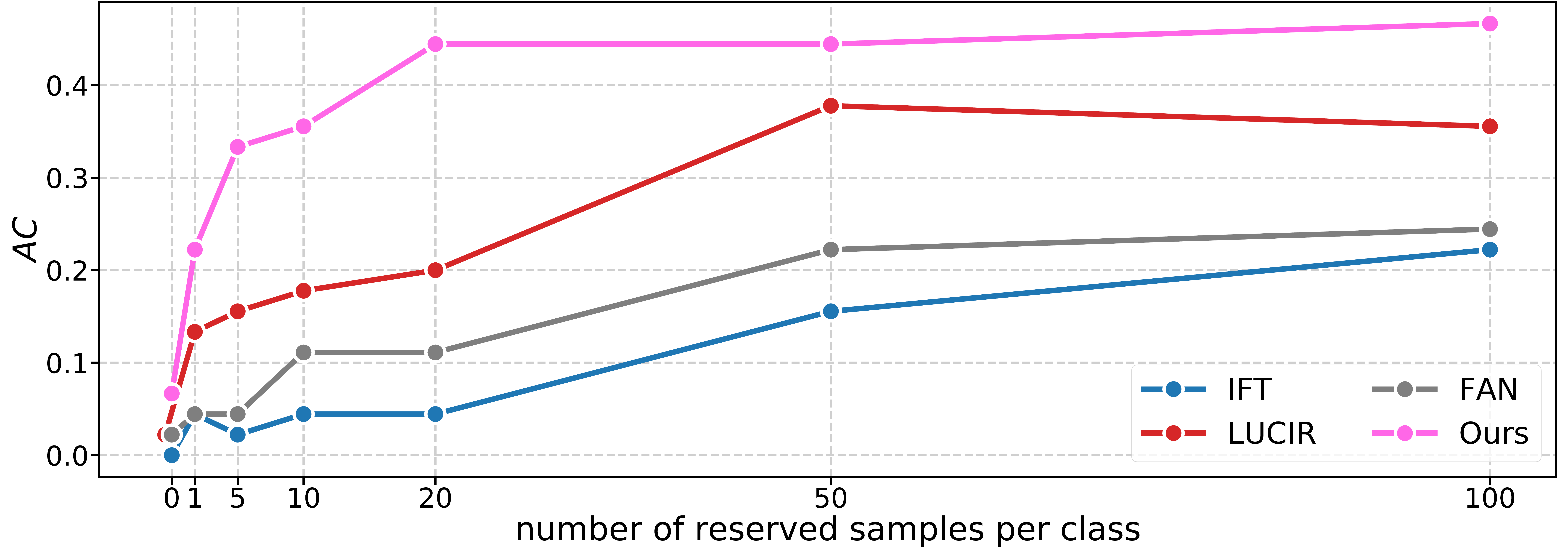}
    \caption{The effect of the number of reserved samples per class in the episodic memory.}
    \label{fig:memory_ablation}
\end{figure}
\paragraph{\textbf{Impact of the Episodic Memory Size.}}
Fig.~\ref{fig:memory_ablation} shows the effect of different numbers of reserved samples per class for both our learning procedure and other baselines. 
As expected, the more samples per class are reserved in the episodic memory, the better the performance. Our approach, with 20 samples per class, achieves results similar to those obtained by the other methods with more examples per class. 
Although ER, LUCIR and FAN have a better relative improvement with 50 samples per class, overall our approach results in the highest performance in learning compatible features.

We also evaluated the methods in the challenging memory-free training setting, (i.e., without the episodic memory). Our training procedure achieves the highest results also in this condition remarking on the fact that CiL methods typically do not have an inherent mechanism to learn compatible features. 

\section{Conclusions} \label{sec:concl}

In this paper, we have introduced the problem of Compatible Lifelong Learning Representations (\clr) that considers the compatibility learning problem within the lifelong learning paradigm. We introduced a novel set of metrics to properly evaluate this problem and proposed a novel \clr training procedure that imposes global and local stationarity on the learned features to
achieve compatibility between representations under catastrophic forgetting. Global and local stationarity is imposed according to the $d$-Simplex fixed classifier and the feature distillation loss, respectively. Empirical evaluation of the learned lifelong-compatible representation shows the effectiveness of our method with respect to baselines and state-of-the-art methods.

\paragraph*{\textbf{Acknowledgments}.}
This work was supported by the European Commission under European Horizon 2020 Programme, grant number 951911 - AI4Media.
The  authors also acknowledge  the  CINECA award  under the ISCRA initiative (ISCRA-C - ``ILCoRe'', ID:~HP10CRMI87), for the availability of high performance computing resources.

\bibliographystyle{ACM-Reference-Format}
\bibliography{bibliography}

\end{document}

%% file: algorithm/cl2r_procedure.tex
\begin{algorithm}[t]
    \caption{\textsc{ \clr Training Procedure}}
    \label{alg:cl2r_train}
    \begin{algorithmic}
        \Require $\{ (\mathcal{D}_t, K_t) \} _{t=1}^T$  
        \Require Model $\phi_t$ with parameters $\Theta$        \Comment{Current model parameters}
        
        \State $\mathcal{M}_1 \gets \emptyset$ 
            \State Initialize the $d$-Simplex fixed classifier $\mathbf{W}$ 
            \State Set $\mathbf{W}{.\mathtt{requires\_grad=False}}$  \Comment{Exclude from training}
        \For{each task $t = \{1, 2, ..., T \}$}
            \State $\mathcal{T}_t \gets \mathcal{M}_{t} \cup \mathcal{D}_t$  \Comment{Current training-set of task $t$}
            \State \textbf{Update} the model $\sigma(\phi_t \circ \mathbf{W})$ on the training-set $\mathcal{T}_t$
            \Indent
                \State $\mathcal{L}_t(\Theta)\big |_{\mathcal{T}_{t}} + \lambda \cdot \lfdmformula (\Theta)\big |_{\mathcal{M}_{t}} \gets $ SGD
            \EndIndent
            \State $\mathcal{M}_{t+1} \gets \mathcal{M}_{t} \cup {\rm \textsc{Sampling}}(D_{t})$
            \Comment{Update memory with sampled data from $\mathcal{D}_t$ }
        \EndFor
    
    \end{algorithmic}
\end{algorithm}

\begin{algorithm}[t]
    \caption{\textsc{Compatible \clr Search} }
    \label{alg:cl2r_search}
    \hspace{-20pt}
    \begin{algorithmic}
        \Require $\phi_{t}$ \Comment{Model under learning by Alg.~1}
        \Require $F_\mathcal{G}=\{\mathbf{f}_i\}_{i=1}^{N_{\mathcal{G}}}$ 
        \Comment{Gallery-set \textit{features}} 
        \Require $I_\mathcal{Q}=\{\mathbf{x}_i\}_{i=1}^{N_{\mathcal{Q}}}$ \Comment{Query \textit{images}}
        \State $F_\mathcal{Q}$$\gets \phi_{t}(I_\mathcal{Q})$ \Comment{Extract query-set feature vectors}
         \State \textsc{Search}$(F_\mathcal{Q}, F_\mathcal{G})$ \Comment{Compatible Search}
    \end{algorithmic}
\end{algorithm}

%% file: tables/datasets.tex
\begin{table}[t]
\centering 

\caption{
Datasets used in \clr training procedures. Training-set and test-set of the same configuration have non overlapping classes to properly evaluate different approaches in a open-set setup.
} \label{tab:datasets} 

\setlength{\tabcolsep}{10pt}

\begin{tabular}{lccccc} 
\toprule 
& & \multicolumn{2}{c}{ \shortstack{\textsc{training-set}} } & \multicolumn{2}{c}{ \shortstack{\textsc{test-set}} }
\\

\cmidrule(lr){3-4} 
\cmidrule(lr){5-6} 

  \shortstack{\textsc{network}} 
& \shortstack{\textsc{input size}}
& \shortstack{\textsc{dataset}} 
& \shortstack{\textsc{\# classes}} 
& \shortstack{\textsc{dataset}} 
& \shortstack{\textsc{\# pairs}} 
\\

\midrule
ResNet-32   &  $32 \times 32$   & CIFAR100       &   100  &   CIFAR10   & 6k \\
ResNet-18   &  $64 \times 64$   & Tiny-ImageNet200  &   200   &   ImageNet20 & 6k \\
ResNet-18   &  $224 \times 224$ & ImageNet900    &   900   &   ImageNet100  & 6k \\
ResNet-50   &  $112 \times 112$ & CASIA-WebFace  &   10575   &   LFW        & 6k \\
ResNet-50   &  $112 \times 112$ & CASIA-WebFace  &   10575   &   IJB-C        & 15M \\
\bottomrule
\end{tabular}

\end{table}

%% file: tables/cifar/2task_cl2r_baseline.tex
\begin{table}[t]
\centering

\caption{
CIFAR10 evaluation. Two-task \clr setting with models trained on CIFAR100.
Initial Task (i.e., the previous task)  shows the verification accuracy on the first 50 classes, the other rows represent the performance obtained after two tasks.
} \label{tab:baselinecomp}

\includegraphics[width=0.7\linewidth]{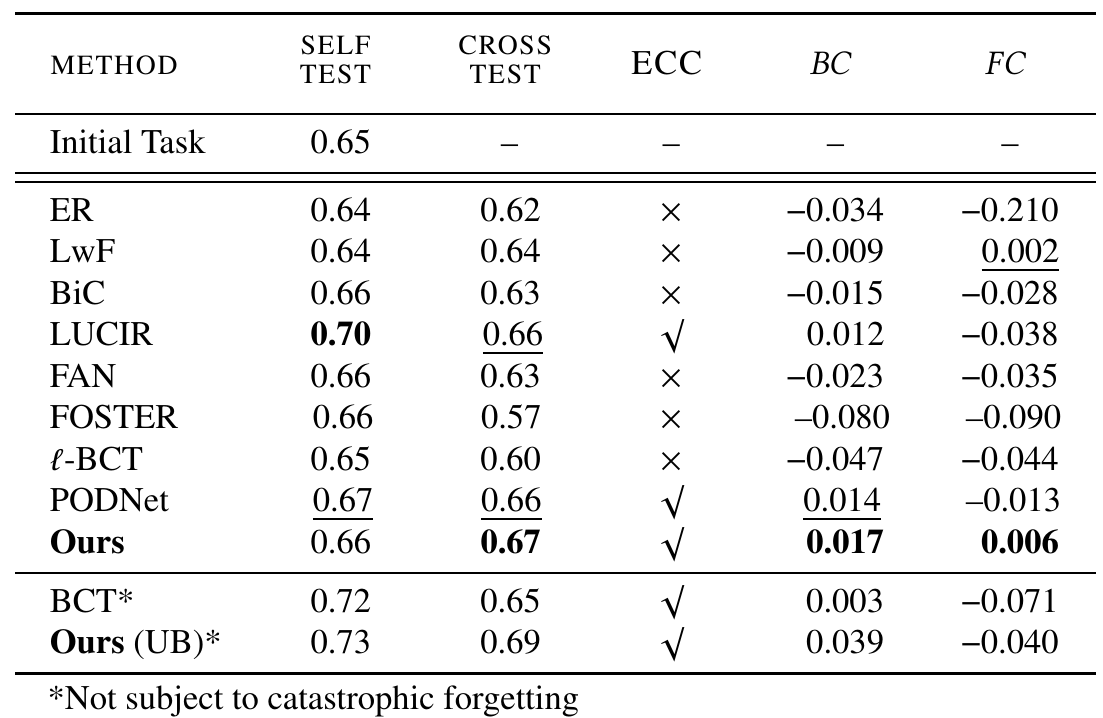}
\end{table}

%% file: tables/cifar/3task.tex
\begin{table*}
\caption{
Evaluation of CIFAR10. Three, five, and ten-task \clr setting with models trained on CIFAR100.
We report \normnecc (Eq. 4), \bc (Eq. 5), and \fc (Eq. 6) for the methods we evaluated. 
} \label{tab:3step} 

\centering 
        
\includegraphics[width=0.9\linewidth]{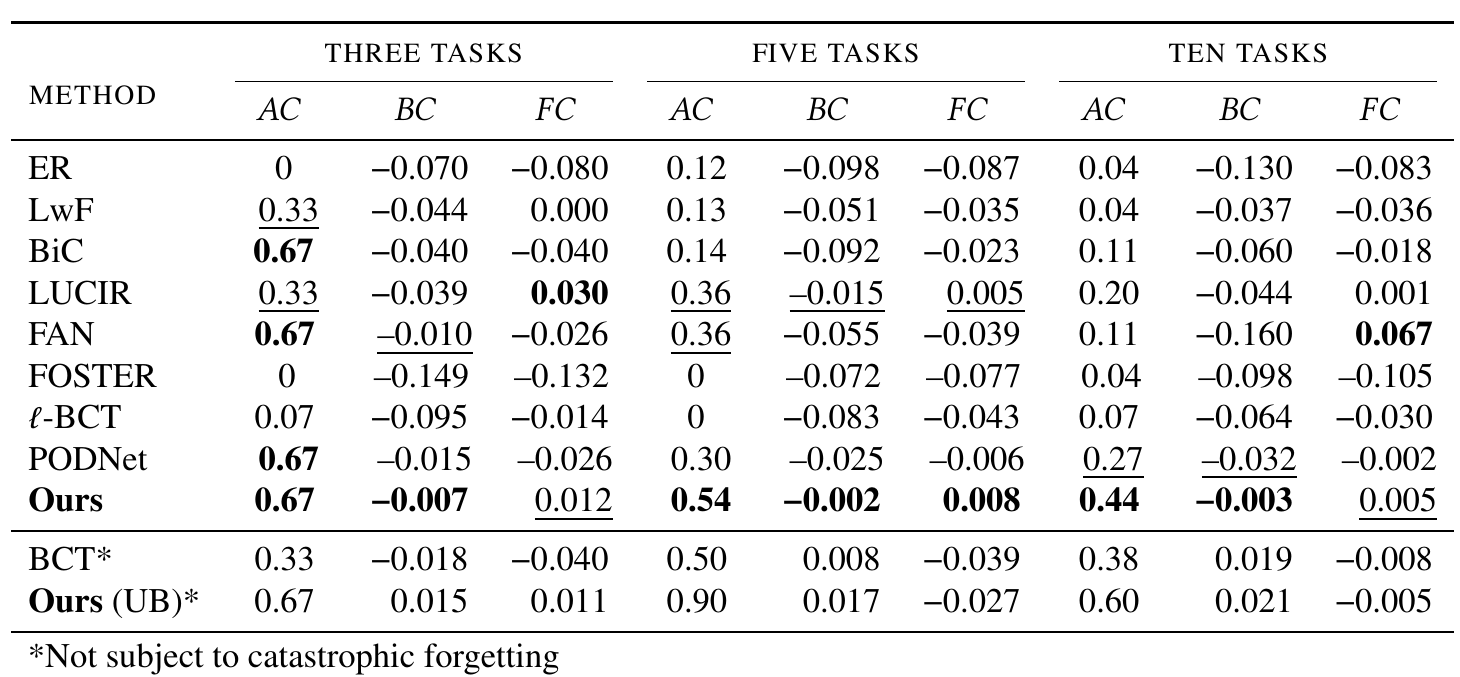}
\end{table*}

%% file: tables/tiny/2task.tex
\begin{table}[t]

\centering 

\caption{
ImageNet20 evaluation. Two-task \clr setting with models trained on Tiny-ImageNet200.
The Initial Task (i.e., the previous task) shows verification accuracy on the first 100 classes, the other rows represent the performance obtained after two tasks.
} \label{tab:two_tiny}

\includegraphics[width=0.7\linewidth]{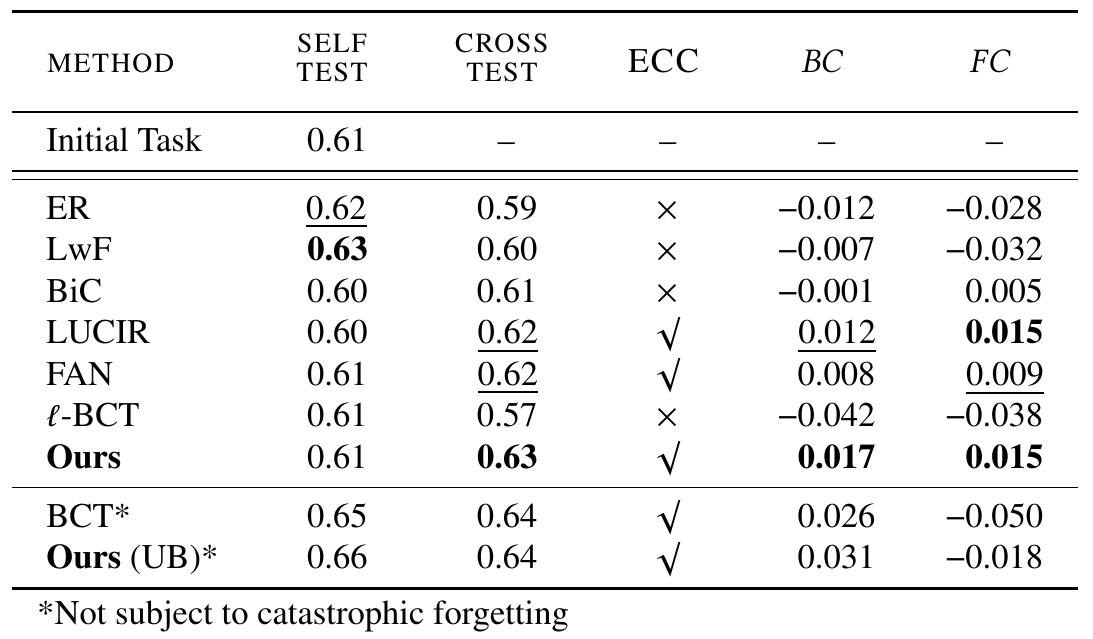}

\end{table}

%% file: tables/tiny/multi_task.tex
\begin{table*}

\caption{
ImageNet20 evaluation. Three, five, and ten-task \clr setting with models trained on Tiny-ImageNet200.
We summarize \normnecc (Eq.~\ref{eq:norm_mecc}), \bc (Eq.~\ref{eq:back_comp}), and \fc (Eq.~\ref{eq:forw_comp}) for the methods we evaluated.
} \label{tab:multi_tiny} 

\centering 
        
\includegraphics[width=0.9\linewidth]{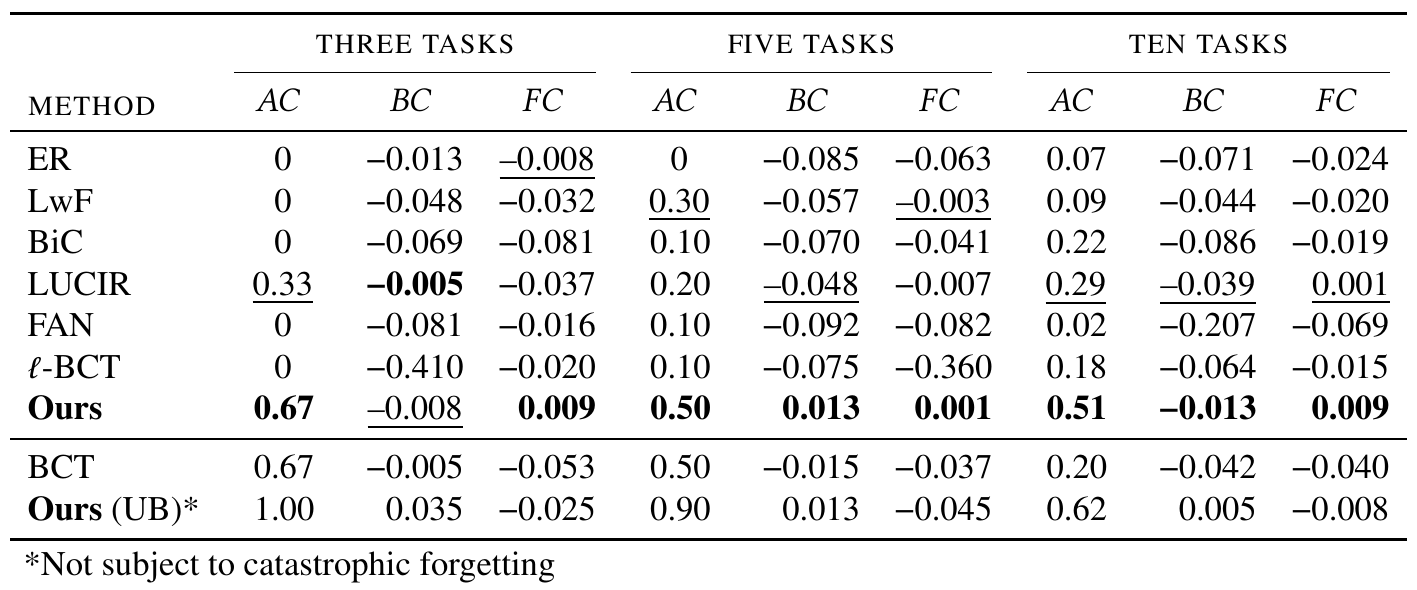}
\end{table*}

%% file: tables/tiny/imagenet.tex
\begin{table*}[t]
\caption{ {
ImageNet100 evaluation. Two and three-task \clr setting with models trained on ImageNet900.
We compare our training procedure and $\ell$-BCT reporting \normnecc, \bc, and \fc. 
}} \label{tab:imagenet} 

\centering 

\includegraphics[width=0.65\linewidth]{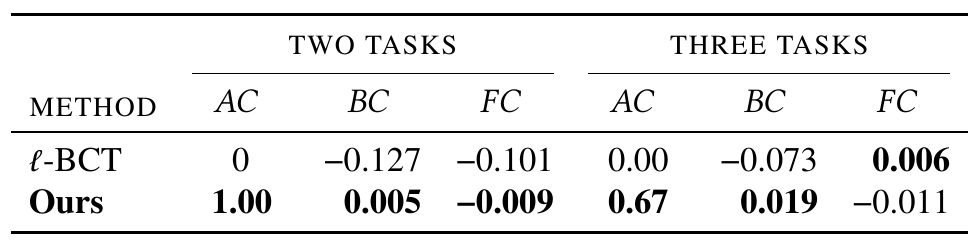}

\end{table*}

%% file: tables/face/multi_task.tex
\begin{table*}[t]
\small
   
\caption{ 
Face verification on LFW dataset. Two, three, five, and ten-task \clr setting with models trained on CASIA-WebFace.
We compare our training procedure and $\ell$-BCT reporting \bc (Eq.~\ref{eq:back_comp}), \fc (Eq.~\ref{eq:forw_comp}), and \normnecc (Eq.~\ref{eq:norm_mecc}), which corresponds to the ECC (Eq.~\ref{eq:compatible_set}) when evaluated in two tasks. 
}\label{tab:face}

\centering 
        
 \includegraphics[width=1.\linewidth]{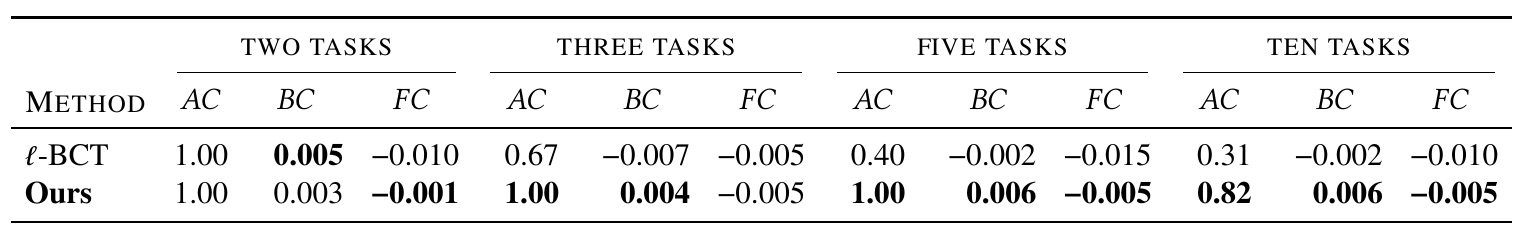}
\end{table*}

%% file: tables/face/ijbc.tex
\begin{table*}[t]
\small
   
\caption{ 
Face verification on IJB-C dataset. Two, three, five, and ten-task \clr setting with models trained on CASIA-WebFace.
We report $\text{(\normnecc, \bc, \fc)@FAR}=\text{10}^{\text{-1}}, \text{ 10}^{\text{-2}}\text{, and }{\text{10}^{\text{-4}}}$ and compare our training procedure with $\ell$-BCT.
}\label{tab:face_ijbc-all} 

\centering

\includegraphics[width=1.\linewidth]{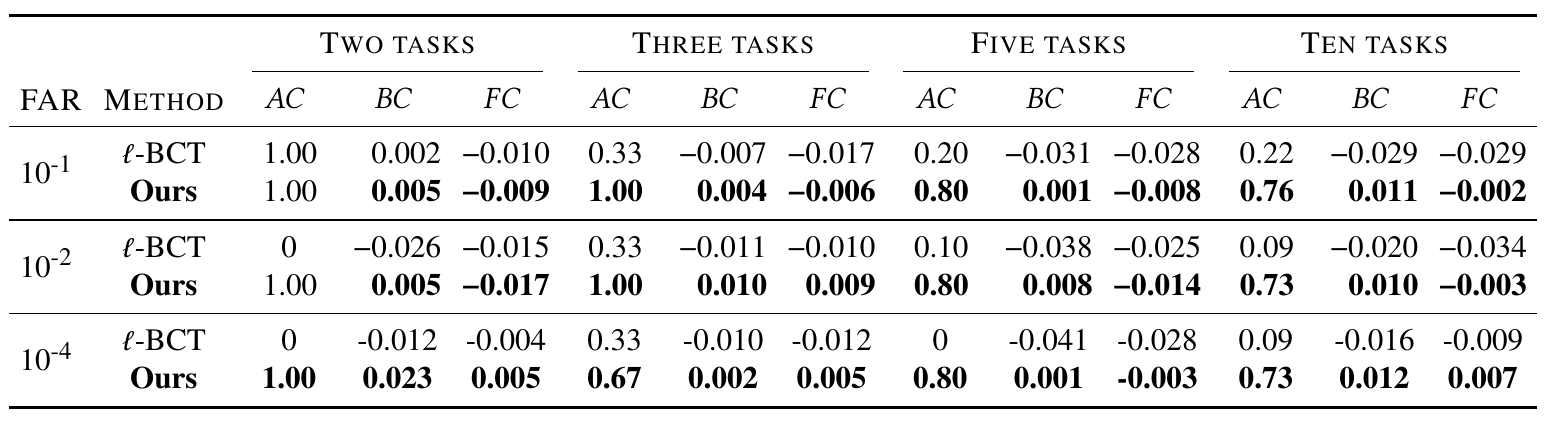}
\end{table*}

%% file: figures/bc_ca_per_task.tex
\begin{figure*}[t]
    \centering
    \subfigure[Five-task \clr setting CIFAR10.]{ 
        \label{fig:bc_per_task_5_cif}
        \adjincludegraphics[width=0.48\textwidth,trim={0 {.1\height} 0 0},clip]{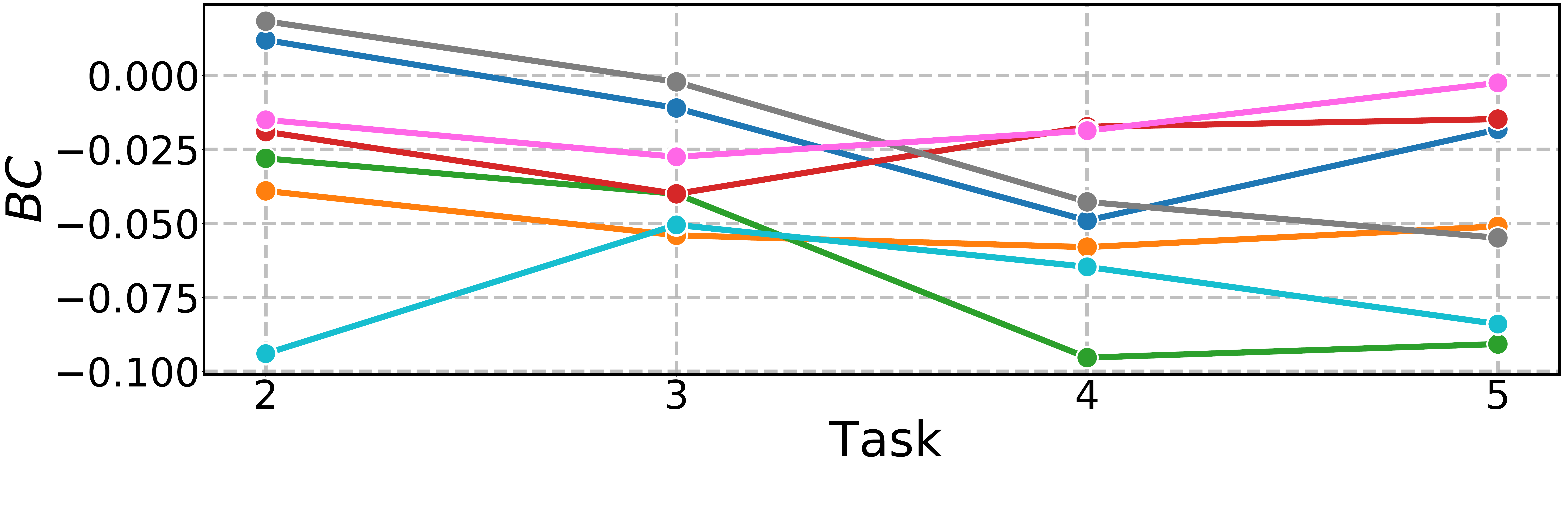}
    }
    \subfigure[Ten-task \clr setting CIFAR10.]{
        \label{fig:bc_per_task_10_cif}
        \adjincludegraphics[width=0.48\textwidth,trim={0 {.1\height} 0 0},clip]{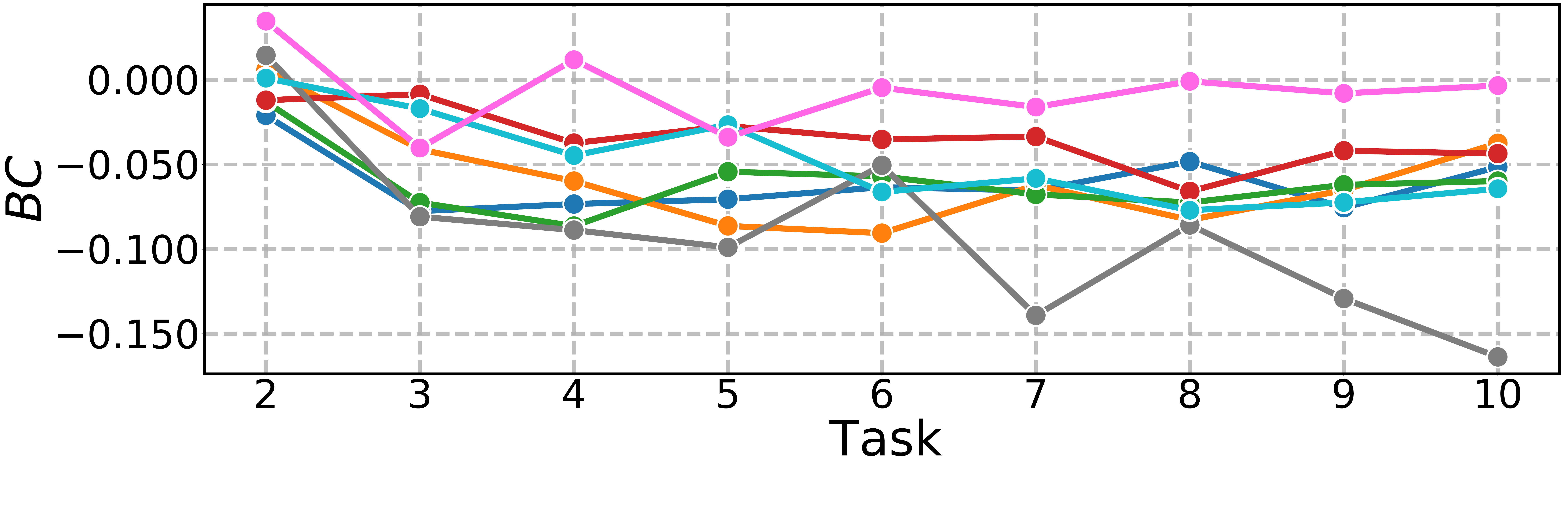}

    }
    \subfigure[Five-task \clr setting ImageNet20.]{ 
        \label{fig:bc_per_task_5_tiny}
        \adjincludegraphics[width=0.48\textwidth,trim={0 {.12\height} 0 0},clip]{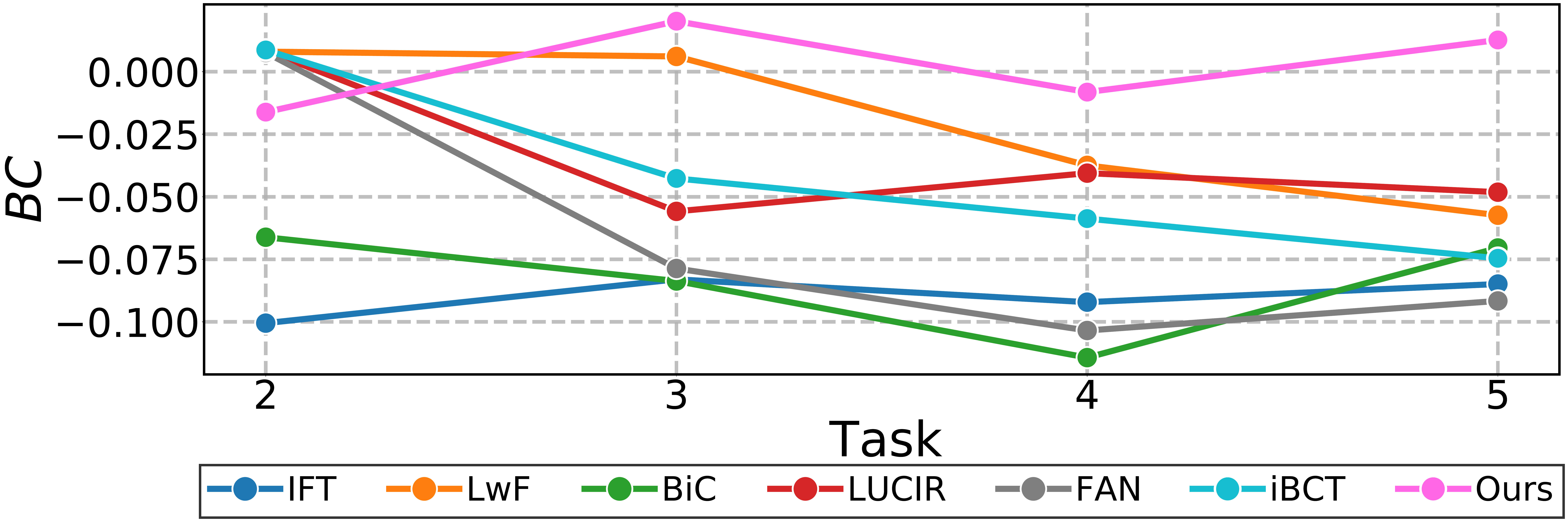}
    }
    \subfigure[Ten-task \clr setting ImageNet20.]{
        \label{fig:bc_per_task_10_tiny}
        \adjincludegraphics[width=0.48\textwidth,trim={0 {.12\height} 0 0},clip]{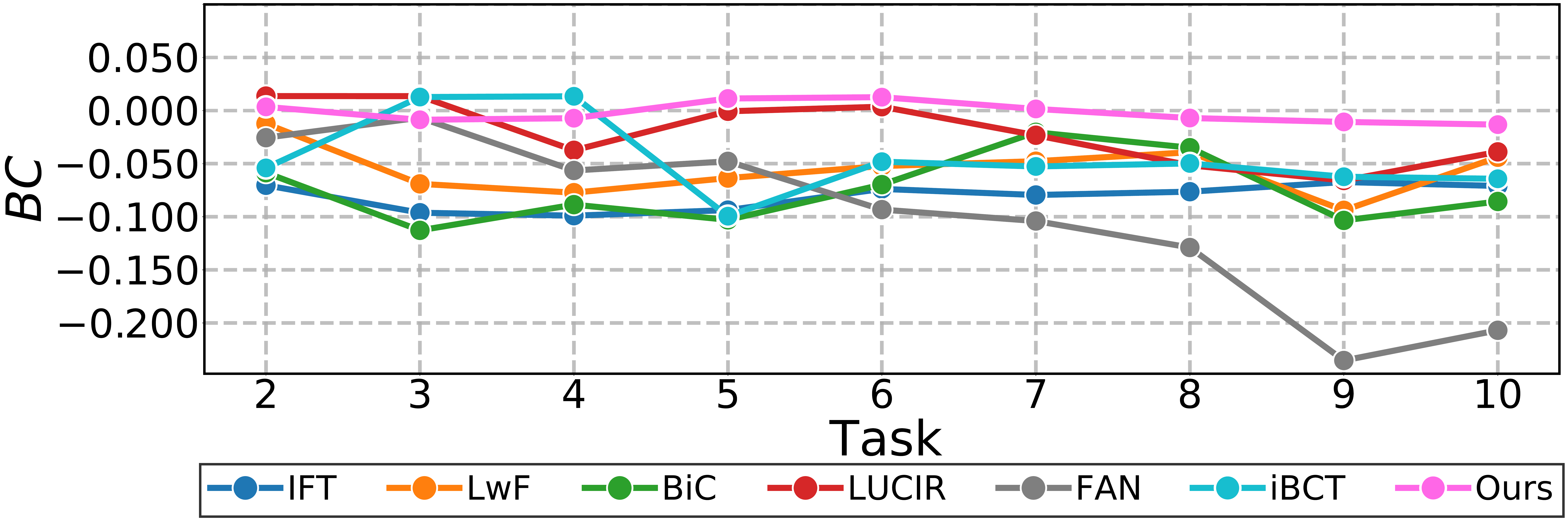}
    }
    \subfigure{
        \nonumber
        \addtocounter{subfigure}{-1}
            \includegraphics[width=0.55\linewidth]{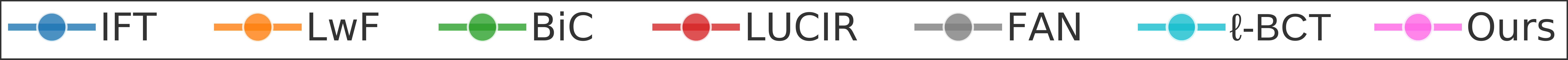}
    }
    \caption{
    Backward Compatibility evolution across tasks $t$ (i.e., $BC(t)$).
    Comparison between our \clr training and other methods in five and ten learning setup. \textit{(a)} and \textit{(b)} CIFAR10 results, \textit{(c)} and \textit{(d)} ImageNet20.
    }
    \label{fig:metric_per_task}
\end{figure*}

%% file: tables/cifar/ablation_repo-fd.tex
\begin{table}[t]
\caption{Ablation of the different main components of our \clr training procedure. The evaluation is performed on CIFAR10 and training is based on CIFAR100 with ten tasks, where: Trainable indicates the traditional Experience Replay (ER) baseline, Fixed indicates ER with stationary features learned from Eq.~\ref{eq:loss_ce} according to the Fixed $d$-Simplex classifier, {\lfdt} is the traditional Feature Distillation, and {\lfdm} is the Feature Distillation evaluated on the only samples stored in episodic memory as defined in Eq.~\ref{eq:feat_dist_on_mem}.
} \label{tab:abl_repo-fd} 

\centering 

\includegraphics[width=0.6\linewidth]{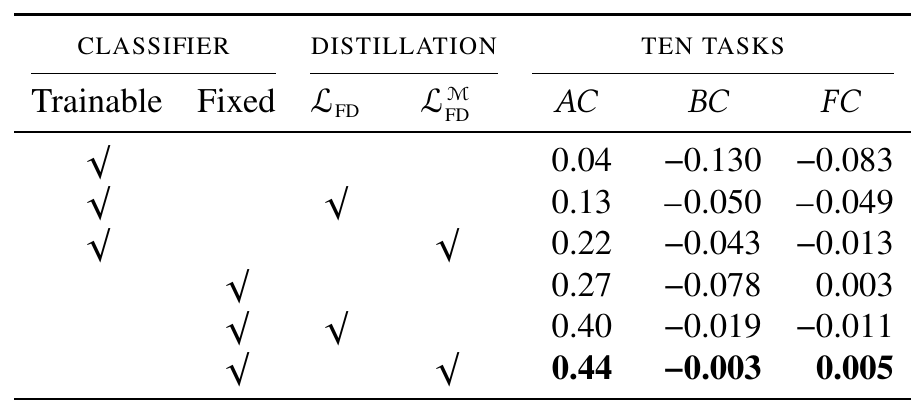}
\end{table}